\def\year{2020}\relax
\documentclass[letterpaper]{article} 
\usepackage{aaai19}  
\usepackage{times}  
\usepackage{helvet}  
\usepackage{courier}  
\usepackage{url}  
\usepackage{graphicx}  
\usepackage{natbib}
\usepackage{hyperref}       
\usepackage[hyperpageref]{backref}

\usepackage{amsmath}
\usepackage{IEEEtrantools}

\DeclareRobustCommand{\citeext}[1]{\citep{#1}}

\begin{document}

\title{Improvement of a dedicated model for open domain persona-aware dialogue generation}
\author{Qiang Han \\
{\tt \{gvvvv\}@163.com} \\}

\maketitle
\begin{abstract}
This paper analyzes some speed and performance improvement methods of Transformer architecture in recent years, mainly its application in dedicated model training. The dedicated model studied here refers to the open domain persona-aware dialogue generation model, and the dataset is multi turn short dialogue, The total length of a single input sequence is no more than 105 tokens. Therefore, many improvements in the architecture and attention mechanism of transformer architecture for long sequence processing are not discussed in this paper. The source code of the experiments has been open sourced\footnote{\url{https://github.com/ghosthamlet/persona}}. 
\end{abstract}

\section[Background]{Background} 
The revolution in the field of NLP (Natural Language Processing) started with the foundation of attention mechanism\citeext {Bahdanau2015}, ~Transformer architecture\citeext{Vaswani2017} ignited the fuse, and officially opened the revolution curtain by the BERT model\citeext{Devlin2019}. It replaced the relatively complex RNN Architecture series models that used to be the mainstream of NLP with simple pure attention mechanism architecture in the past four years, it has changed the field of NLP in an all-round way, and created the Imagenet moment\citeext{ruder2018nlpimagenet} in NLP field. 

Open domain dialogue generation is an important and very complex task of NLP. Transformer architecture has also made a qualitative leap in this task, such as GPT-2, GPT-3 model and Google Meena\citeext{Radford2019, Brown2020, Adiwardana2020}. If there is no special explanation in the following text, the models involved are Transformer architecture models. 

GPT-2 and GPT-3 models are general models, which are suitable for almost all NLP tasks, and are not dedicated to open domain dialog generation. These models usually need huge model size, massive pretraining data and long pretraining time to match the performance of dedicated medium or small models (although DistilGPT2\footnote{\url{https://github.com/huggingface/transformers/tree/master/examples/distillation}} and DistilBERT\citeext{Sanh2019} are used to optimize the general model, but their performance also decreases accordingly). 
Moreover, it is difficult to generate a conversation with consistent personality\footnote{The chatbot has specific and fixed personality characteristics, such as name, gender, address, etc.} in the general model, even if method like TransferTransfo\citeext{Wolf2019} which finetuning GPT-2 model\footnote{Adds personalized data to the input dialog context}, although it has good performance on the personalized dialogue English dataset PERSONAL-CHAT\citeext{Zhang2018}, it is still not as good as the dedicated model\citeext{Zheng2019} when trained on personalized Dialogue Chinese dataset PersonalDialog\citeext{Zheng2019a}. GPT-3 only has beta API, no need to fine tune, using prompt programming by prepending the personalized data, but we have not obtained the test authority, unable to compare the performance. 

In this paper, we introduced some speed and performance improvement methods of Transformer architecture in recent years to this dedicated open domain persona-aware dialogue generation model\citeext{Zheng2019}, and analyze the effective and invalid attempts on this dedicated model. In the following text, open domain persona-aware dialogue generation is referred to as persona-aware dialogue generation. 

\section[Related Works]{Related Works} 
The research on persona-aware dialogue generation is less than that of no persona-aware dialogue generation and closed domain task-oriented dialogue generation, so the relevant dialogue datasets are also less. The English ones are mainly PERSONA-CHAT\citeext{Zhang2018}, while the Chinese ones are PersonalDialog\citeext{Zheng2019a} and Personality Assignment\citeext{Qian2017}. Most of the studies discussed here mainly use these three datasets. 

The early persona-aware dialogue generation is basically the Seq2Seq model\citeext{SutskeverGoogle2014} of RNN+attention mechanism\citeext{Bahdanau2015}, such as the complex multi-stage training of\citeext{Qian2017} (dataset Personality Assignment), simple end-to-end training of\citeext{Zheng2019a} (dataset PersonalDialog) etc., as well as the Seq2Seq model of RNN+memory mechanism\citeext{Sukhbaatar2015}, such as several baseline models in\citeext{Zhang2018} (dataset PERSONA-CHAT). 

With the rise of Transformer architecture, recently most of the models have been the original general-purpose or special-purpose models of Transformer architecture, such as\citeext{Tselousov2018} (dataset PERSONA-CHAT) changing the model of GPT\citeext{Radford2018}, \citeext{Wolf2019} (dataset PERSONA-CHAT) with the native GPT, and \citeext{Zheng2019} (dataset PersonalDialog) adding Attention Routing, and \citeext{Liu2020} (dataset PERSONA-CHAT) with combination of GPT and BERT, and \citeext{Roller2020} (dataset is mainly PERSONA-CHAT, with another three) with changed Transformer and so on. 

In addition, there are relatively few models using VAE (Variational AutoEncoder), RL (Reinforcement Learning) and GAN (Generative Adversarial Network), such as CVAE (Conditional VAE)+RNN+memory mechanism\citeext{Song2019} (dataset PERSONA-CHAT), VAE+GRU(RNN)+memory mechanism+attention mechanism\citeext{Xu2020} (dataset PERSONA-CHAT), RL+Transformer\citeext{Liu2020} (dataset PERSONA-CHAT) and so on, We have not studied the GAN related models and will not introduce them here. 

\section[Model]{Model} 

We choose the Attention Routing model of Transformer architecture\citeext{Zheng2019} as the research object, because of its simplicity and efficiency, and does not deviate from the original Transformer architecture too far, so we can use most of the improvement of native Transformer architecture. In addition, the dataset we used is PersonalDialog, but due to limited resources, we can't use its complete millions of data. Training and evaluation are only conducted on 100000 and 20000 sessions randomly extracted without replacement. The training and evaluation of the full dataset will be studied in the future. The Attention Routing model is called AR model, and our improved model is called AR+ model. 

The AR model is an Encoder Decoder structure, similar to the full version of the Transformer architecture\citeext{Vaswani2017}, with the following differences: 

1. In terms of input representation, the dialog context uses~\_SEP special characters as separator and spliced into a sequence. The sequence is segmented as characters, did not use BPE or SentencePiece. Context embedding and persona embedding of corresponding speakers are summing together, and then input into Encoder. The persona key-value pairs of target are spliced into another sequence and input into the same Encoder after embedded. 

2. In terms of model architecture, the Encoder and the Decoder share weight. The Decoder has only one attention module, that is, the Attention Routing module. The target sequence is input into the Decoder after embedding, and attend to itself to get attention $O_{prev}$, attend to context encode to get attention $O_C$, attend to persona encode to get attention $O_T$, then sum up these three attentions, $O_T$ and $O_C$ is multiplied by a weight respectively, and an additional item of $O_C$ is added: 
\begin{equation}
O_{merge} = aO_T + (1 - a)O_C + O_C + O_{prev}    
\end{equation}

The weight of $a$ is controlled by a supervised dynamic weight predictor subnetwork, which requires additional supervised learning. Therefore, we do not consider it in our study for the moment, and directly set the weight to 1: 
\begin{equation}
O_{merge} = O_T + O_C + O_{prev}    
\end{equation}

3. In the aspect of training, the multitask method is adopted. In addition to the dialogue generation task, language model task is added. Crossentropy is used. Loss is as follows: 
\begin{equation}
L(\phi, \theta) = L_D(\phi) + \lambda_1L_{LM}(\phi) + \lambda_2L_W(\theta)
\end{equation}

$L_D(\phi)$ is dialogue generation loss, $\lambda_1~\lambda_2$ is loss weight hyperparameter, $L_{LM}(\phi)$ is language model loss, $L_W(\theta)$ is the Predictor loss, it's not considered here, so the final loss is:
\begin{equation}
L(\phi) = L_D(\phi) + \lambda_1L_{LM}(\phi)
\end{equation}

For a more detailed description, please refer to the original paper\citeext{Zheng2019}. \\

On the basis of retaining the overall structure of AR model, AR+ model introduces the following improvements: 

1. ReZero\citeext{Bachlechner2020} method, a simple architecture change of gating each residual connection using a single zero-initialized parameter, and removed all norms except the pre norm. Accelerated the model convergence. There are two differences between the treatment of AR+ model and the original paper of ReZero. A. the original paper does not retain pre norm, B. AR+ add a fix attention to attention $O_T~O_C$ at the residual junction: 
\begin{equation}
O_{output} = E_{prev} + bO_T + bO_C + d(O_{merge}) * r
\end{equation}

$O_{output}$ is output of the residual connection, $E_{prev}$ is prev output, $b$ is the fix attention hyperparameter, default to 0.1, $d(*)$ is dropout, $r$ is zero-initialized learnable parameter.

2. ALBERT\citeext{Lan2019} method, factoring the embedding layer, reducing the embedding dimension and making it independent of the hidden dimension of the model, so the two can be modified separately; share the weight of the transformer layer. These two modifications reduce the amount of calculation, model size and memory consumption. 

3. Factor FF method, factorize the two fully connected layers within the transformer layer. Reduce the amount of calculation, model size and memory consumption. 

4. MemN2N\citeext{Sukhbaatar2015} method, there is no order of the target speaker persona key-value pairs, and its embedding is simple. No need to use Transformer to encode. Instead, we uses word segmentation and memory mechanism to process it, result in better performance, and reduced the amount of calculation. 

5. BART MLM\citeext{Lewis2019} method, BART paper shows that mask language model is better than autoregressive language model in most cases, so we use mask language model task similar to BART in multi task training. 

\section[Experiments]{Experiments} 
AR+ model hyperparameters:

Characters vocabulary size: 9489, embedding size: 200, embedding pretrained on full PersonalDialog\citeext{Zheng2019a} datasets, word vocabulary size of persona: 10004, persona embedding size is same as model hidden size: 512, 
Transformer layers: 6, attention head: 8, FF layer size: 2048, dropout: 0.1. 
Batch size: 64, epoch: 3, optimizer: AdamW (Bias corrected AdamW from Transformers library\footnote{\url{https://github.com/huggingface/transformers/}}), we use lr finder\citeext{Smith2015} (library\footnote{\url{https://github.com/davidtvs/pytorch-lr-finder}}) to find lr should be: 0.2e-2, weight decay: 0.05, clip grad: 1.
Language model $L_{LM}$ $\lambda_1$: 0.5, lr scheduler: ReduceLROnPlateau, scheduler params: mode min, factor 0.5, min\_lr 1.5e-4, patience 60.
MemN2N params: hops 3, layer\_share adjacent.

Baseline model is AR model, lr: 1.5e-4, other hyperparameters is same as AR+. As many people have said, the learning rate is indeed the most influential of all the hyperparameters. If the AR model use the same 0.2e-2 learning rate as AR+, then the AR model can not learn anything at all, and may be that the training data subset we used is not large enough, which makes the first epoch overfitted. 

The decoding strategy of baseline and AR+ is Nucleus Sampling\citeext{Holtzman2019}, params: temperature 0.7, top\_k 0, top\_p 0.9.

\subsection[Datasets]{Datasets} 
The original data of PersonalDialog\citeext{Zheng2019a} dataset contains some duplicate data. After deduplication, there are 5,195,149 sessions. The training, validation and test datasets are 100000, 20000 and 20000 session subsets which are extracted randomly without replacement after data deduplication. We use the validation set to adjust the hyperparameters and evaluate on the test set. 

Dialogues are generally short sentences with an average length of 15 characters. Therefore, the length of dialogue is limited to a maximum of 15 characters, and the context is limited to a maximum of three turns, together with the question, the maximum context length is $15*(2*3+1)=105$ characters. The length of target is also limited to 15 characters. Persona includes gender, address, interests, and interests can include multiple items. 

\subsection[Training]{Training}
We study the optimization effect of different methods, so the pretraining is not included in the experiment unless otherwise specified. 

\subsection[Evaluation]{Evaluation} 
We only analyze the automatic metric methods, including: 1. BLEU\citeext{Papineni2002}, 2. F1, 3. PPL(Perplexity), 4. Dist.(Distinct)\citeext{Li2016}, as well as training speed, memory consumption and model size. The manual evaluation is reserved for future study. 

\subsection[Result]{Result} 
AR+ model is better than AR model in most metrics. Dist1 and Dist2 have little difference, which may be due to the small training dataset and insufficient training. See the table for specific data: \ref{tab:result-comparisons}.

\begin{table*} [b]
\centering
\caption{Experimental results, bold data is improved metric}
\resizebox{2\columnwidth}{!}{%
\begin{tabular}{|l || c|c|c|c|c || c|c|c |}
\hline
\textbf{Model} & BLEU & F1 & PPL 
    & Dist1 & Dist2 & Params & GPU mem & train time \\
\hline
AR  & 0.00257 & 0.00017 & 614 
    & 0.842 & 0.772 & 30M & 7920M & 30.5m \\
AR+ & \textbf{0.00510} & \textbf{0.00024} & \textbf{120} 
    & 0.830 & 0.780 & 31M & \textbf{5600M} & \textbf{21.0m} \\
\hline
\end{tabular}%
}
\label{tab:result-comparisons}
\end{table*}

\subsection[Ineffective Methods]{Ineffective Methods} 
The following is an analysis of ineffective Methods: 

1. Adapters\citeext{Houlsby2019}, an alternative lightweight fine-tuning strategy. They consist of a small set of additional newly initialized weights at every layer of the transformer. These weights are then trained during fine-tuning, while the pre-trained parameters of the large model are kept frozen/fixed. Maybe because the AR+ model is different from the transformer or the pretraining data and time are not enough, the loss of the adapter method is too large to complete the fine-tuning. 

2. Use Pretrained features\citeext{Devlin2019}, the embedding part of AR+ model is replaced by the features of pretrained ALBERT\citeext{Lan2019} or ELECTRA\citeext{Clark2020}. The improvement of loss is very limited or even worse. We also test feature+embedding, feature replace encoder, features of different layers, combination of features of different layers, and pretrained models of different sizes, all the results are similar. 

3. Use Pretrained weights to init AR+\citeext{Ziegler2019}, the weight of AR+ model is initialized with the weights of pretrained ALBERT or ELECTRA, the improvement of loss is limited, sometimes worse. We also try to cancel the sharing of Encoder and Decoder, let Encoder or Decoder initialize separately, cancel layer sharing, initialize the corresponding layer of pretrained model, and pretrained model of different sizes, all the results are not different. 

Methods 2 and 3 were validated in the papers \citeext{Devlin2019} and \citeext{Zhao2019} respectively, but they could not produce an effect on AR+ model. We also tested the pretrained model of BERT or XLNET\citeext{Yang2019} and the effect was not much changed. We suspect that A. These pretrained models are all BERT branch (except XLNET), which are suitable for various tasks such as classification, question answering and regression, But not suitable for transferring to the task of text generation, especially the task of dialogue generation in the open domain. Although \citeext{Zhao2019} has successfully used Bert in the dialogue model, their dialogue model is a special VAE+RNN model, b. The training data of these pretrained models are ordinary webpage or article content, which is quite different from the fine-tuning dialogue data, so the positive effect can be very limited. 

The pretrained models are all from Transformers library\citeext{Wolf2019HuggingFacesTS}.

In addition, in some cases, pretraining maybe not necessarily necessary, let's look at the original purpose of pretraining: 

1. After pretraining, the general model can be used to fine tune different tasks to achieve the purpose of reuse.

2. When the labeled dataset is not large enough, the pretraining can improve the performance and shorten the fine tuning time.

If the labeled dataset is large enough to reach more than 5 million, and the model is a dedicated model. Then if there is a ready-made pretrained general model, just try to use it. But if there is no ready-made suitable pretrained model, it is unnecessary to pretrain the dedicated model. First of all, the purpose of 1 cannot be achieved, and then the purpose of 2 may not be achieved. After all, there are large-scale labeled data, trained from scratch in the same or even shorter time as pretraining+fine tuning, may yield similar performance.

\section[Ablation Study]{Ablation Study} 
We removed each of the five optimizations for ablation study, and compared with the complete AR+ model, we can see that these optimizations are respectively in the improvement of training speed, the reduction of model size, the reduction of memory consumption and the improvement of performance have produced significant effects. After ablation, the data of corresponding metrics have changed accordingly. See the table for specific data: \ref{tab:result-ablation}.
 
\begin{table*} [b]
\centering
\caption{Ablation study results, bold data is degradation metric}
\resizebox{2\columnwidth}{!}{%
\begin{tabular}{|l || c|c|c|c|c || c|c|c |}
\hline
\textbf{Model} & BLEU & F1 & PPL 
    & Dist1 & Dist2 & Params & GPU mem & Train time \\
\hline
AR+ & 0.00510 & 0.00024 & 120 
    & 0.830 & 0.780 & 31M & 5600M & 21.0m \\
-ReZero & \textbf{0.00371} & \textbf{0.00019} & \textbf{172} 
        & 0.821 & 0.746 & 31M & 5360M & 19.5m \\
-ALBERT & \textbf{0.00473} & 0.00027 & \textbf{123} 
          & 0.850 & 0.784 & \textbf{45M} & \textbf{5890M} & 20.4m \\
-Factor\_FF & \textbf{0.00345} & 0.00024 & 110 
             & 0.875 & 0.789 & \textbf{33M} & \textbf{6690M} & \textbf{24.6m} \\
-MemN2N & \textbf{0.00260} & \textbf{0.00021} & 117 
          & 0.878 & 0.807 & 9M & \textbf{7410M} & \textbf{25.3m} \\
-BART\_MLM & \textbf{0.00499} & 0.00024 & \textbf{636} 
            & 0.831 & 0.780 & 31M & 5120M & 16.7m \\
\hline
\end{tabular}%
}
\label{tab:result-ablation}
\end{table*}

\section[Conclusion]{Conclusion} 
In this paper, we improve the dedicated persona-aware dialogue generation model with recent advances in Transformer architecture. Experiments show that even if the model is transformer architecture, as long as the internal structure is changed, the improvements for the original architecture may not be suitable for the new model. In addition, the pretrained BERT branch models are not suitable for transfer to the dialogue generation transformer model. Since we only use small models to train the subsets of PersonalDialog dataset, in the future research we can scale the model to train the full dataset to test whether the five effective methods proposed in this paper are still effective in large-scale training. 


\bibliographystyle{aaai}

\end{document}